\documentclass[sigconf]{acmart}

\usepackage{caption}
\usepackage{amsmath}
\usepackage{subcaption}
\usepackage{multirow}
\usepackage{multicol}
\usepackage{color}
\usepackage{colortbl}
\usepackage{diagbox}
\newcommand{\tableCellHeight}{1}
\newcommand{\tabstyle}[1]{
  \setlength{\tabcolsep}{#1}
  \renewcommand{\arraystretch}{\tableCellHeight}
  \centering
  \small
}
\definecolor{mygray}{gray}{.9} 

\newcommand{\rotbox}[1]{\rotatebox{55}{#1}}
\AtBeginDocument{%
  }

\acmISBN{978-1-4503-XXXX-X/2018/06}

\renewcommand\footnotetextcopyrightpermission[1]{}

\begin{document}

\title{AttriPrompt: Dynamic Prompt Composition Learning for CLIP}


\author{Qiqi Zhan}
\orcid{0009-0008-9061-5979}
\email{zhanqiqi@buaa.edu.cn}
\affiliation{%
  \institution{Beihang University}
  \city{Beijing}
  \country{China}
}

\author{Shiwei Li}
\orcid{0000-0002-8765-9283}
\email{shiweili93@buaa.edu.cn}
\affiliation{%
  \institution{Hangzhou Innovation Institute,\\Beihang University}
  \city{Hangzhou}
  \country{China}
}

\author{Qingjie Liu}
\orcid{0000-0002-5181-6451}
\authornote{Corresponding authors}
\email{qingjie.liu@buaa.edu.cn}
\affiliation{%
\institution{Beihang University}
\city{Beijing}
  \country{China}
}

\author{Yunhong Wang}
\orcid{0000-0001-8001-2703}
\email{yhwang@buaa.edu.cn}
\affiliation{%
  \institution{Beihang University}
  \city{Beijing}
  \country{China}
}


\begin{abstract}
The evolution of prompt learning methodologies has driven exploration of deeper prompt designs to enhance model performance. However, current deep text prompting approaches suffer from two critical limitations: Over-reliance on constrastive learning objectives that prioritize high-level semantic alignment, neglecting fine-grained feature optimization; Static prompts across all input categories, preventing content-aware adaptation. To address these limitations, we propose AttriPrompt-a novel framework that enhances and refines textual semantic representations by leveraging the intermediate-layer features of CLIP’s vision encoder. We designed an Attribute Retrieval module that first clusters visual features from each layer. The aggregated visual features retrieve semantically similar prompts from a prompt pool, which are then concatenated to the input of every layer in the text encoder. Leveraging hierarchical visual information embedded in prompted text features, we introduce Dual-stream Contrastive Learning to realize fine-grained alignment. Furthermore, we introduce a Self-Regularization mechanism by applying explicit regularization constraints between the prompted and non-prompted text features to prevent overfitting on limited training data. Extensive experiments across three benchmarks demonstrate AttriPrompt's superiority over state-of-the-art methods, achieving up to 7.37\% improvement in the base-to-novel setting. The observed strength of our method in cross-domain knowledge transfer positions vision-language pre-trained models as more viable solutions for real-world implementation.


\end{abstract}




\maketitle

\section{Introduction}

\begin{figure}
	\centering
	\includegraphics[width=0.9\columnwidth]{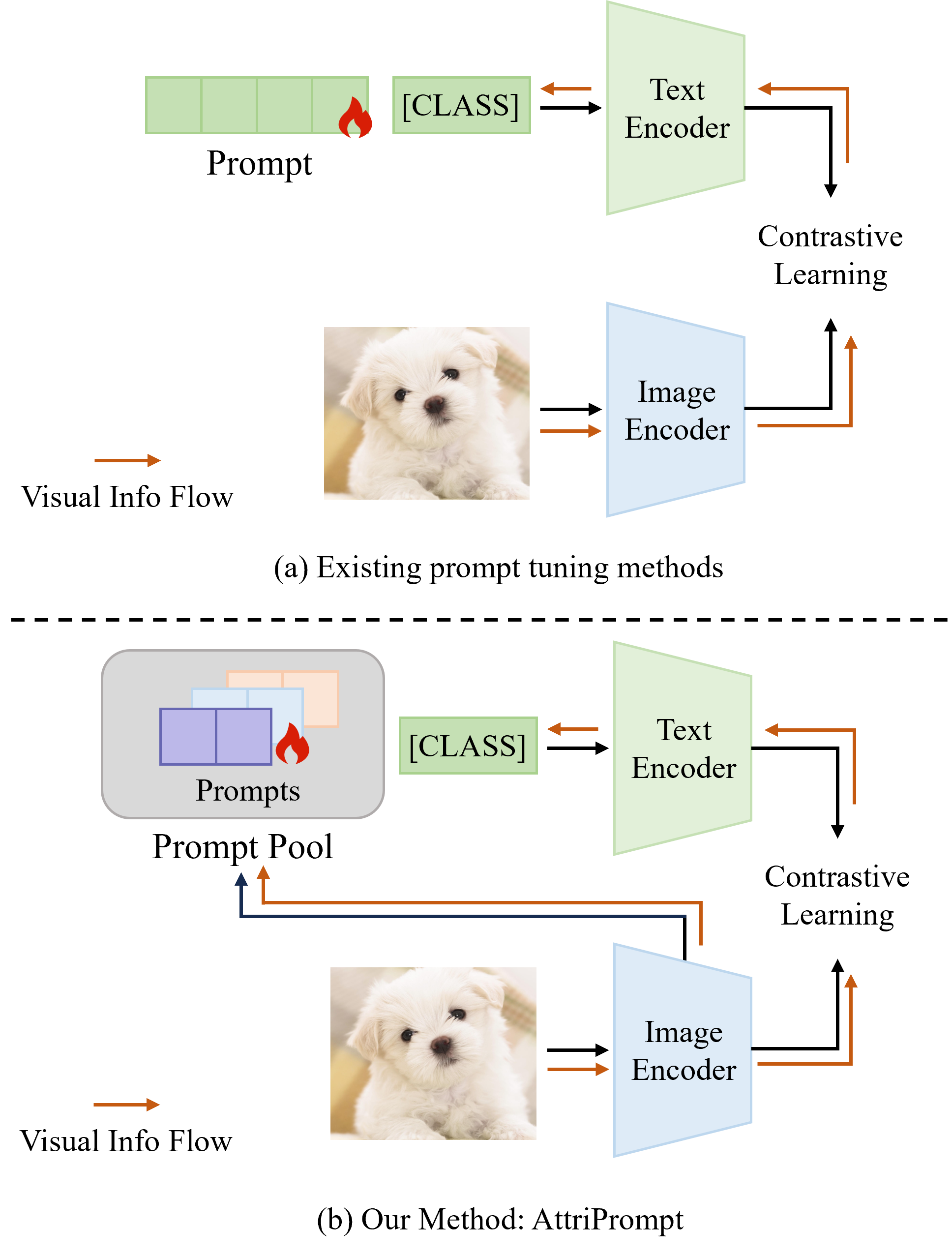} 
	\caption{Comparison of our method with other prompt learning methods. (a) Existing methods suffer from a single propagation path of visual information and lack the ability to dynamically adjust the text prompt based on the visual input. (b) Our method introduces a novel visual information propagation path, which facilitates the refinement of text features and enhances cross-modal alignment. Meanwhile, the proposed dynamic prompt composition enables content-aware adaptation, thereby improving the model's generalization. }
	\label{fig:intro}
\end{figure}

Recent years have witnessed remarkable advances in large vision-language pre-trained models (VLPMs) ~\cite{jia2021scaling,radford2021learning,yao2021filip,yuan2021florence,zhai2022lit}. These models are trained on web-scale datasets using contrastive loss, allowing them to align image-text pairs within a shared embedding space and encode open-vocabulary concepts. This training paradigm equips VLPMs with both task-specific knowledge and cross-modal association capability, enabling inference on unseen categories or concepts through learning visual-linguistic correlations. Models such as CLIP ~\cite{radford2021learning} and ALIGN ~\cite{jia2021scaling} therefore offer robust support for downstream tasks, including open-vocabulary image recognition~\cite{kim2021adapt, zhai2023sigmoid}, object detection~\cite{feng2022promptdet}, and image segmentation~\cite{luddecke2022image, DBLP:conf/iclr/LiWBKR22}. Despite acquiring broad conceptual representations during training, these model struggle to generalize effectively when faced with new tasks or categories that deviate substantially from the training data. In such scenarios, the model’s zero-shot generalization performance may deteriorate severely.

Prompt learning, inspired by the success of prompt engineering in NLP~\cite{ding2021openprompt,radford2019language}, optimizes a small set of learnable prompt vectors using a few training samples from the target task while keeping the pretrained model weights frozen. However, this approach may result in overfitting to the specific task’s data distribution, consequently diminishing the generalization capability of VLPMs. Recent studies have investigated various strategies to mitigate the performance degradation of fine-tuned models on novel tasks. Nevertheless, balancing task adaptation and generalization remains a critical research challenge.

Existing prompt learning methods show a trend of development from shallow to deep layers. Starting with the CoOP~\cite{zhou2022learning} method, which adds learnable embedding vectors only at the input layer, to the deeper-level PromptSRC~\cite{khattak2023self} method that introduces learnable embedding at higher layers, researchers have been exploring ways to improve model performance through deeper prompt vectors. However, these deep text prompt-based methods have certain limitations. Such methods often "fail to see" the specific content of the current input image, instead optimizing text prompts exclusively through contrastive learning objectives. Due to these limitations, text-prompt-based optimization faces several significant challenges: Current frameworks transmit visual information to textual modalities through contrastive learning objectives, which focus exclusively on high-level semantic alignment. Thus, low-level visual features are not effectively leveraged, potentially hindering the model from capturing essential visual cues required for the task. Moreover, existing methods cannot dynamically adjust the text prompt based on the content of the image. This prevents the model from adapting flexibly to various image, thereby restricting its generalization across multi-domain tasks.

Human visual recognition of novel objects follows a hierarchical pathway: initial detection of basic feature attributes (e.g., colors, stripes, and outlines), followed by activation of semantic associations from memory, culminating in recognition by integrating these information.
Building on this insight, we propose AttriPrompt, which fully leverages the intermediate-layer features of CLIP's vision encoder to optimize and refine textual semantic representations. In order to transmit fine-grained visual information into the text encoder while enabling adaptive prompt adjustment, we design an Attribute Retrieval module. This module first clusters visual features from different layers, mirroring the human visual feature recognition process. Next, these aggregated attribute features retrieve the most similar semantic prompt from the prompt pool, as illustrated in Figure \ref{fig:intro}. Finally, the retrieved learnable prompts are concatenated into the input of each text encoder layer, and these prompted features through the CLIP framework produce the final prediction.

Given that the prompts incorporate hierarchical visual information, we introduce a Dual-stream Contrastive Learning, encouraging the model to focus on the retrieved learnable prompts. This refine alignment mechanism improves model's performance by facilitating the integration of visual information into the cross-modal representation. While leveraging CLIP's pretrained knowledge for inference tasks, models constrained by data scarcity risk catastrophic forgetting through overfitting to downstream tasks. To mitigate this, we introduce Self-Regularization and apply explicit regularization constraints, i.e. a L1 regularization strategy between the prompted text features and the non-prompted text features. This approach effectively guides the model's optimization process while mitigating overfitting to the specific task data.

Overall, the contributions of our work can be summarized as:
\begin{itemize} \setlength{\itemsep}{2pt}
\item We propose AttriPrompt, a method inspired by our study of human cognitive processes. We designed the Attribute Retrieval module to simulate human cognition, transmitting additional visual information into the text prompt, enhancing the cross-modal alignment ability. At the same time, it enables adaptive feature selection, improving generalization to unseen categories. 

\item We propose a Self-Regularization Mechanism, which helps guide the model’s optimization process, and a Dual-stream Contrastive Learning mechanism, which helps refining cross-modal alignment, to further improve the generalization capability. 

\item Our method demonstrates outstanding performance in base-to-novel class generalization, domain generalization, and cross-dataset settings, proving its effectiveness and robustness. 


\end{itemize}


\section{Related Works}

\subsection{Vision Language Models}

Vision-language pretraining models~\cite{jia2021scaling,radford2021learning,yao2021filip,yuan2021florence,zhai2022lit} act as a bridge to facilitate the unified understanding of visual and textual data by utilizing large-scale internet data for pretraining. CLIP~\cite{radford2021learning}, a prominent vision-language pretraining model, was pretrained on approximately 400 million image-text pairs, employing contrastive learning as its core training objective. Contrastive learning~\cite{chen2020simple} brings similar image-text features closer while separating non-matching pairs, thereby aligning images and texts within a shared embedding space. This process allows CLIP to effectively encode the rich semantics of an open vocabulary in a unified embedding space and capture the underlying cross-modal relationships. 

Due to this distinctive training strategy, CLIP excels in a variety of tasks, especially in challenging tasks, including few-shot classification, zero-shot classification, and generalization~\cite{ding2022decoupling,gao2024clip,gu2021open,maaz2022class,manzoor2023multimodality,rao2022denseclip,bangalath2022bridging,zang2022open}, demonstrating exceptional performance. However, adapting these foundational models to specific downstream tasks while optimizing their performance without compromising their original generalization capabilities remains a significant challenge.  Given that the data available for downstream tasks is typically much smaller than the large-scale datasets used during pretraining, models encounter a substantial data scarcity issue, which can significantly reduce their generalization ability for specific tasks. Therefore, effectively leveraging the potential of pretrained models with limited data, while avoiding overfitting, remains a critical challenge in current research.

\subsection{Prompt Learning}

Prompt tuning is among the most widely used methods for adapting vision-language pre-trained models to downstream tasks. This approach stems from the field of NLP~\cite{ding2021openprompt,radford2019language}, where manually designed templates guide the model in generating outputs relevant to specific downstream tasks. However, due to the high cost of manually configuring these templates, recent research has shifted focus to soft prompts, a transfer learning method that involves adding a small number of learnable prompts to the input. CoOP~\cite{zhou2022learning} is the first method to apply prompt tuning to a vision-language pre-trained model such as CLIP. It fine-tunes the model by incorporating a set of continuous, learnable prompt vectors into the language branch. However, this method may result in overfitting on task-specific or domain-specific data. CoCoOP~\cite{zhou2022conditional} employs a meta-learning approach, generating task-specific vectors for each sample to mitigate the overfitting issue. Bahng et al.~\cite{bahng2022visual} conduct visual prompt tuning by learning prompts for the visual branch of CLIP. Bayesian Prompt~\cite{derakhshani2023bayesian} models the input prompt space from the perspective of probability distribution, treating it as a prior distribution, making it compatible with unconditional or image-conditioned prompt learning methods. ProDA~\cite{zhu2023prompt} introduce a data-driven method which can learn low-bias prompts from a few samples. Previous prompt learning methods have predominantly focused on text-based or image-based approaches, while largely overlooking the fusion and interaction between the two modalities. Later, MaPLe~\cite{khattak2023maple} presents a multimodal prompt tuning method that enhances model transfer and generalization by improving the fusion of prompts from different modalities. PromptSRC~\cite{khattak2023self} enhances the model's generalization capability by leveraging pre-trained features for model regularization.

\section{Preliminaries}

In this section, we first introduce the structure of CLIP and the settings of the learnable prompt.

\subsection{Introduction of CLIP}

CLIP consists of a vision encoder $f$, based on either ResNet~\cite{he2016deep} or ViT~\cite{vaswani2017attention}, and a text encoder $g$,  based on Transformer. The pretraining objective of CLIP is formulated as a contrastive learning loss applied to image-text pairs. The input image is denoted by $X \in \mathbb{R}^{C\times H \times W}$. Before being processed by the vision encoder $f$, the input image is partitioned into $N$ patches, with an additional learnable embedding vector $e_{cls}$ appended. The final feature representation input to the vision encoder is denoted as $\widetilde {\textbf{X}} = \{e_{cls}, e_1, e_2, ..., e_N\}$. These visual patches $\widetilde X$ are encoded by the vision encoder $f$ to extract latent features, deriving a global representation of the image $\widetilde f = f(\widetilde X) \in \mathbb{R}^d$. For textual input, a prompt template, such as "a photo of [CLASS]", is employed, where [CLASS] denotes a specific category label. This can be formulated as $\widetilde {\textbf{Y}}=\{t_{SOS}, t_1, t_2, ...,t_L, t_{class}, t_{EOS}\}$, where ${t_l}$ represents word embeddings, $t_{class}$ corresponds to the class token, and $t_{SOS}$ and $t_{EOS}$ denote the learnable start and end embeddings, respectively. These token embeddings are encoded by the text encoder to extract the latent textual features $\widetilde g = g(\widetilde Y) \in \mathbb{R}^d$. The inference process of CLIP can be express as:

\begin{equation}
\mathcal{P}(\hat y=i \textbar X) = \frac{\exp(\cos(\widetilde g_i, \widetilde f/\tau))}{\sum_{j=1}^C \exp(\cos(\widetilde g_j, \widetilde f / \tau))}
\label{eq:clip-infere}
\end{equation}
where $ \mathcal{P}(\hat y=i \textbar X)$ represents the probability of the input $X$ belonging to class $i$, $C$ is the number of class, $\cos(\cdot, \cdot)$ denotes the cosine similarity and $\tau$ is the temperature.

\subsection{Prompt Learning for CLIP}

The Prompt-based CLIP adaption method, first introduced by CoOp , involves learning soft prompts to enhance the transfer performance on downstream tasks. Specifically, CoOp appends $T$ learnable word embeddings, which are shared across all classes, denoted as $\textbf{P}_t = \{p_t^1,p_t^2,...,p_t^T\}$. Therefore, the text encoder processes the input token sequence $\widetilde {\textbf{Y}}_p = \{t_{SOS}, \textbf{P}_t, t_1, t_2, ...,t_L,t_{EOS}\}$ to generate prompted text feature represented as $\widetilde g_p = g(\widetilde Y_p)$. 
For an image-text pair $(X, Y)$ from the dataset $D$, the image feature $\widetilde f = f(\widetilde X)$ is encoded by the vision encoder $f$, and the text feature $\widetilde g_p = g(\widetilde {Y_p})$ is obtained from the text encoder $g$. The model's parameters are optimized with the following cross-entropy loss:

\begin{equation}
\mathcal{L}_{ce} = \text{argmin}\ \mathbb{E}_{(X,Y)\sim D} \mathcal{L}( \text{sim}(\widetilde f, \widetilde {g_p}), Y)
\label{eq:ce-loss}
\end{equation}
where $\mathcal{L}$ represents cross-entropy loss function.

\section{Methodology}

In this section, we first provide an overview of our works. Then, the three core steps of the method are discussed, these being the Attributes Retrieval, Self-Regularization and Dual-stream Contrastive Learning.

\subsection{Overview}

\begin{figure*}
    \centering
    \includegraphics[width=1.0\linewidth]{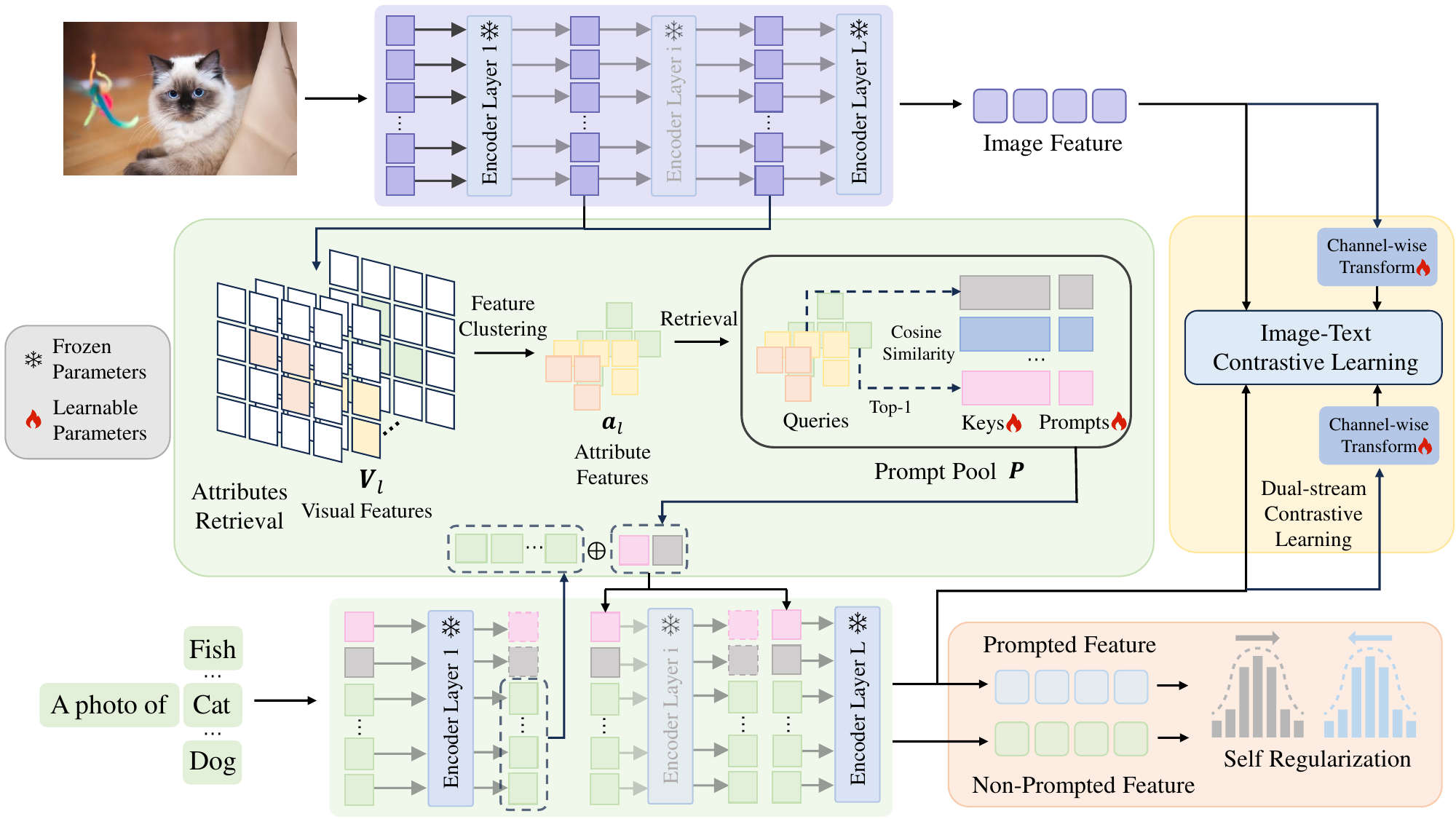}
    \caption{The framework of AttriPrompt, which is composed of Attributes Retrieval, Self Regularization and Dual-stream Contrastive Learning. The Attribute Retrieval module performs per-layer feature clustering in visual encoder. The clustered attribute features retrieve the most similar prompts from the prompt pool based on cosine similarity. These retrieved prompts are then concatenated with the corresponding layer’s text embeddings. We propose a Dual-stream Contrastive Learning, enhancing the model's ability to exploit fine-grained alignment. To alleviate the problem of overfitting, a L1 loss between the prompted features and the non-prompted features is introduced to guide the model’s training process.}
    \label{fig:overview}
\end{figure*}

Figure \ref{fig:overview} provides an overview of our methods. Our method utilizes fine-grained visual features extracted from the intermediate layers of CLIP’s vision encoder. The proposed Attribute Retrieval module clusters the visual features extracted from intermediate layers and retrieves the most relevant prompts from a prompt pool. The retrieved prompts are then concatenated with the text features at the corresponding layers. Next, to enhance the utilization of CLIP’s inherent generalization capability, we propose a Self-Regularization between the prompted text features and the non-prompted text features. This approach enables the model to adapt to downstream data while preserving and incorporating CLIP’s generalization. Meanwhile, to further refine the alignment between the visual and textual modalities, a Dual-stream Contrastive Learning is introduced.

\subsection{Attributes Retrieval}

For a given input image $X$, the vision encoder $f$ segments the image into non-overlapping patches and processes these patches using a series of stacked Transformer blocks to extract a set of visual feature representations $\textbf{V}_L = \{v_{L_0}, v_{L_1}, v_{L_2}, ..., v_{L_N}\}$, where $N$ represents the number of patches and $L$ represents the number of the model layers. For each layer $l$ of the vision encoder, a set of vision features $\textbf{V}_l$ is extracted. These features encode specialized visual characteristics, such as edges, textures, and distinct conceptual categories. Therefore, we propose to leverage these internal local features in model as representations of visual attributes for categories. This approach extracts and employs fine-grained visual feature information from the model itself, instead of depending on external resources or auxiliary knowledge for visual attribute description. 

Specifically, we construct a prompt pool $\textbf{P}$, which consists of $M$ learnable prompts, each with a length of $L_p$. Additionally, each learnable prompt $p_i$ is paired with a key vector $k_i$ for retrieval, which shares the same dimension as the image feature. For the visual features $\textbf{V}_l$ from a specific layer $l$, we first cluster the vision tokens into $k$ attribute features $a_l = \{a_{l_1}, a_{l_2}, ..., a_{l_k} \}$. Subsequently, the attribute features retrieve the most similar learnable semantic embeddings from the prompt pool $\textbf{P}$. Given clustered attribute features, we calculate the cosine similarity between the attribute features and key, followed by applying a softmax function to obtain probabilities, as follows:

\begin{equation}
S_{i,j} = \frac{\exp (\cos(a_{l_i}, k_j))}{\sum_{j=1}^M \exp(\cos(a_{l_i}, k_j))} 
\label{eq:retrival}
\end{equation}

Next, we choose the keys with the highest cosine similarity score to form a set of prompts $ \textbf{P}_l = \{p_{l_i}\}_{i=1}^{k}$, where $p_{l_i}$ denotes the $i$-th prompt chosen uniquely for $a_{l_i}$. Subsequently, for the $l$-th layer encoder, the retrieved prompts are concatenated to the corresponding text features to serve as additional feature information injected into the text representations. The input embeddings of the $l$-th layer now follow the form $[p_{l_1}, p_{l_2}, ..,p_{l_k}, \textbf{W}_l]$, where $ \textbf{W}_l = [w^1_l, w^2_l, ..., w^N_l]$ are the fixed tokens. The new learnable tokens are progressively incorporated into each transformer block of the language encoder up to a specified depth.

\begin{equation}
[\_, \textbf{W}_i] = g_i([\textbf{P}_{i-1}, \textbf{W}_{i-1}]), i=1,2,...,N_g
\label{eq:deep_prompt}
\end{equation}
where $g_i$ represents the $i$-th layer of text encoder, $N_g$ denotes the number of layers in the text encoder, and $[\cdot ,\cdot]$ represents the concatenation operation.


\subsection{Self-Regularization}

Although we froze CLIP's pretrained weights to prevent degradation of generalization capability, these learnable prompts gradually adapted to the specific data distribution of downstream tasks, contrary to our original design objective of having prompts guide the model to focus on task-agnostic yet semantically meaningful features. Therefore, we introduce a Self-Regularization, by adding explicit constraint loss, to preserve the model’s generalization ability.

For a given input sample, text features are obtained through learnable prompts and pre-trained text features, $\widetilde g_p$ and $\widetilde g$. We impose a constraint on both the prompted and non-prompted text features to maintain consistency with the CLIP-pretrained features, as follows:

\begin{equation}
\mathcal{L}_{cc} = \sum_{i=1}^d \vert \widetilde g_p - \widetilde g \vert
\label{eq:cc-loss}
\end{equation}
where $d$ denotes the dimension of text features. Here, we follow \cite{khattak2023self} and adopt the L1 loss as the consistency constraint for the two types of text features. By adding this loss, the model is able to learn task-agnostic but invariance knowledge from downstream data by adjusting its learning direction according to the original frozen CLIP model.

\subsection{Dual-stream Contrastive Learning}

As CLIP does not leverage intermediate-layer visual representations during its pre-training phase, direct alignment between image and text features may be the optimal choice. Therefore, we introduce an additional module for refining alignment, which activates channel-wise features that are otherwise neglected by CLIP, encouraging the model to attend to the newly introduced information. Specifically, we add two channel-wise transformation heads at the output of both the vision encoder and the text encoder, which map text and image features to new feature dimensions. Formally, $\textbf{V}^{'}_L = \{v^{'}_0, v^{'}_1, v^{'}_2, ..., v^{'}_N\}$, and

\begin{equation}
v^{'}_i = \alpha \odot v_i + \beta, i=1,...,N
\label{eq:transform}
\end{equation}
where $\alpha$ and $\beta$ are the trainable scaling and shift vectors. The text features are subjected to the same transformation. The transformed image features $\widetilde f^{'}$ and text features $\widetilde g^{'}$ are aligned in the new feature space using CLIP's loss function, enabling the model to focus on newly introduced information. For each pair of $(X,Y)$ from dataset $D$, the secondary alignment minimizes the following loss:

\begin{equation}
\mathcal{L}_{align} = \text{argmin}\ \mathbb{E}_{(X,Y)\sim D} \mathcal{L}(\text{sim}(\widetilde f^{'}, \widetilde g^{'}), Y)
\label{eq:align-loss}
\end{equation}
where $\mathcal{L}$ represents cross-entropy loss function.

\subsection{Training and Inference}

In the training phase, we utilize the supervised contrastive loss, denoted as $L_{ce}$. According to Equation \ref{eq:clip-infere} and \ref{eq:ce-loss}, the loss can be expressed as:

\begin{equation}
    \mathcal{L}_{ce}=\mathbb{E}_{(X,Y)\sim D}[-\text{log}\frac{\exp({\cos(\widetilde g_i, \widetilde f/\tau)})}{\sum_{j=1}^C \exp({\cos(\widetilde g_j, \widetilde f/\tau)})}].
    \label{eq:Lm}
\end{equation}

In addition, we introduce an attributes retrieval matching loss that brings the most relevant prompt key and image embedding closer together. This loss leverages cosine similarity to measure the proximity, and its formula is as follows:

\begin{equation}
    \mathcal{L}_{match} = \sum_l^{N_f} \sum_i^k \cos(a_{l_i}, k_j)
    \label{eq:match-loss}
\end{equation}
where $N_f$ represents the number of layers in the image encoder. 

Finally, to enhance the semantic diversity of the learned prompts, we introduce an additional loss which orthogonalizes the embeddings of distinct prompts. This helps mitigate semantic redundancy in the prompt vectors, thereby increasing prompt diversity and enabling the model to generalize across a broader range of downstream tasks and data.

\begin{equation}
    \mathcal{L}_{div} = \frac{1}{M(M-1)} \sum_{i=1}^M \sum_{j=i+1}^M \cos(p_i,p_j)
\end{equation}

The overall loss can be formulated as follows:
\begin{align}
    \mathcal{L} =\ &(1-\lambda_1)\mathcal{L}_{ce} + \lambda_1 * \mathcal{L}_{align} \\
    & + \lambda_2*\mathcal{L}_{cc} +  \lambda_3 * \mathcal{L}_{div}  -  \lambda_4* \mathcal{L}_{match} \notag
\end{align}

During the inference phase, the test-time knowledge fusion will be used for prediction. The predictions from the two alignment heads are combined, and the final output prediction is obtained through a weighted fusion approach. The formula is as follows:

\begin{equation}
    \mathcal{P}(c_i \vert x) = (1-\lambda_1) \mathcal{P}_{ce}(c_i \vert x) + \lambda_1 \mathcal{P}_{align}(c_i \vert x)
\end{equation}
where $c_i$ denotes the predicted class, $\mathcal{P}_{ce}$ represents the prediction probability from the original CLIP, and $\mathcal{P}_{align}$ denotes the probability from our additionally introduced head.

\begin{table*}[t!]
\small
\centering

\renewcommand{\arraystretch}{0.82}
{
    \caption{Comparison with state-of-the-art methods on base-to-novel generalization. Our proposed method demonstrates strong generalization results over existing methods on 11 recognition datasets. The best results are in bold and the second-best results are underlined. HM indicates the harmonic mean.}
    \begin{subtable}[t]{.32\textwidth}
    \centering
    \caption{\textbf{Average over 11 datasets.}}
    \begin{tabular}{l cc|c}
    \toprule
    Method & Base & Novel & HM \\
    \midrule
    CLIP & 69.34 & 74.22 & 71.70 \\
    CoOp & 82.69 & 63.22 & 71.66 \\
    CoCoOp & 80.47 & 71.69 & 75.83 \\
    MaPLe & 82.28 & 75.14 & 78.55 \\
    ProGrad & 82.48 & 70.75 & 76.16 \\
    KgCoOp & 80.73 & 73.60 & 77.00 \\
    PromptSRC & \underline{84.26} & \underline{76.10} & \underline{79.97} \\
    \midrule
    \rowcolor{mygray} 
    Ours & \textbf{84.88} & \textbf{77.63} & \textbf{81.09} \\
    \bottomrule
    \end{tabular}
    \end{subtable}
    ~
    \begin{subtable}[t]{.32\textwidth}
    \centering
    \caption{ImageNet.}
    \begin{tabular}{l cc|c}
    \toprule
    Method & Base & Novel & HM \\
    \midrule
    CLIP & 72.43 & 68.14 & 70.22 \\
    CoOp & {76.47} & 67.88 & 71.92\\
    CoCoOp & 75.98 & {70.43} & {73.10} \\
    MaPLe & {76.66} & 70.54 & {73.47} \\
    ProGrad & 77.02 & 66.66 & 71.46 \\
    KgCoOp & 75.83 & 69.96 & 72.78 \\
    PromptSRC & \underline{77.60} & \underline{70.73} & \underline{74.01} \\
    \midrule
    \rowcolor{mygray} 
    Ours & \textbf{77.63} & \textbf{71.00} & \textbf{74.17} \\
    \bottomrule
    \end{tabular}
    \end{subtable}
    ~
    \begin{subtable}[t]{.32\textwidth}
    \centering
    \caption{Caltech101.}
    \begin{tabular}{l cc|c}
    \toprule
    Method & Base & Novel & HM \\
    \midrule
    CLIP & 96.84 & {94.00} & 95.40 \\
    CoOp & {98.00} & 89.81 & 93.73 \\
    CoCoOp & 97.96 & 93.81 & {95.84} \\
    MaPLe & 97.74 & {94.36} & {96.02} \\
    ProGrad & {98.02} & 93.89 & 95.91 \\
    KgCoOp & 97.72 & \underline{94.39} & \underline{96.03} \\
    PromptSRC & \underline{98.10} & 94.03 & 96.02 \\
    \midrule
    \rowcolor{mygray} 
    Ours &  \textbf{98.50} & \textbf{95.53} & \textbf{96.99} \\
    \bottomrule
    \end{tabular}
    \end{subtable}
    
    \begin{subtable}[t]{.32\textwidth}
    \centering
    \caption{OxfordPets.}
    \begin{tabular}{l cc|c}
    \toprule
    Method & Base & Novel & HM \\
    \midrule
    CLIP & 91.17 & 97.26 & 94.12 \\
    CoOp & 93.67 & 95.29 & 94.47 \\
    CoCoOp & {95.20} & {97.69} & {96.43} \\ 
    MaPLe & \underline{95.43} & \underline{97.76} & \underline{96.58}\\
    ProGrad & 95.07 & 97.63 & 96.33 \\
    KgCoOp & 94.65 & \underline{97.76} & 96.18 \\
    PromptSRC & {95.33} &97.30 & 96.30 \\
    \midrule
    \rowcolor{mygray} 
    Ours & \textbf{96.13} &	\textbf{98.07} & \textbf{97.09} \\
    \bottomrule
    \end{tabular}
    \end{subtable}
    ~
    \begin{subtable}[t]{.32\textwidth}
    \centering
    \caption{StanfordCars.}
    \begin{tabular}{l cc|c}
    \toprule
    Method & Base & Novel & HM \\
    \midrule
    CLIP & 63.37 & {74.89} & 68.65 \\
    CoOp & 78.12 & 60.40 & 68.13 \\
    CoCoOp & 70.49 & 73.59 & {72.01} \\
    MaPLe & 72.94 & 74.00 & {73.47} \\
    ProGrad & 77.68 & 68.63 & 72.88 \\
    KgCoOp & 71.76 & \underline{75.04} & 73.36 \\
    PromptSRC & \underline{78.27} & 74.97 & \underline{76.58} \\
    \midrule
    \rowcolor{mygray} 
    Ours & \textbf{80.33} & \textbf{75.50} & \textbf{77.70} \\
    \bottomrule
    \end{tabular}
    \end{subtable}
    ~
    \begin{subtable}[t]{.32\textwidth}
    \centering
    
    \caption{Flowers102.}
    
    \begin{tabular}{l cc|c}
    \toprule
    Method & Base & Novel & HM \\
    \midrule
    CLIP & 72.08 & \textbf{77.80} & 74.83 \\
    CoOp & {97.60} & 59.67 & 74.06 \\
    CoCoOp & 94.87 & 71.75 & {81.71} \\
    MaPLe & 95.92 & 72.46 & {82.56} \\
    ProGrad & 95.54 & 71.87 & 82.03 \\
    KgCoOp & 95.00 & 74.73 & {83.65} \\
    PromptSRC & \underline{98.07} & 76.50 & \underline{85.95} \\
    \midrule
    \rowcolor{mygray} 
    Ours & \textbf{98.33} & \underline{77.40} & \textbf{86.62} \\
    \bottomrule
    \end{tabular}
    \end{subtable}
    
    \begin{subtable}[t]{.32\textwidth}
    \centering
    
    \caption{Food101.}
    
    \begin{tabular}{l cc|c}
    \toprule
    Method & Base & Novel & HM \\
    \midrule
    CLIP & 90.10 & 91.22 & 90.66 \\
    CoOp & 88.33 & 82.26 & 85.19 \\
    CoCoOp & {90.70} & {91.29} & {90.99} \\
    MaPLe & \underline{90.71} & \textbf{92.05} & \textbf{91.38} \\
    ProGrad & 90.37 & 89.59 & 89.98 \\
    KgCoOp & 90.05 & 91.70 & 91.09 \\
    PromptSRC & 90.67 &91.53 &91.10 \\
    \midrule
     \rowcolor{mygray}    
    Ours & \textbf{90.77} & \underline{91.93} & \underline{91.35} \\
    \bottomrule
    \end{tabular}
    \end{subtable}
    ~
    \begin{subtable}[t]{.32\textwidth}
    \centering
    
    \caption{FGVCAircraft.}
    
    \begin{tabular}{l cc|c}
    \toprule
    Method & Base & Novel & HM \\
    \midrule
    CLIP & 27.19 & {36.29} & {31.09} \\
    CoOp & {40.44} & 22.30 & 28.75 \\
    CoCoOp & 33.41 & 23.71 & 27.74 \\
    MaPLe & 37.44 & 35.61 & {36.50} \\
    ProGrad & {40.54} & 27.57 & 32.82 \\
    KgCoOp & 36.21 & 33.55 & 34.83 \\
    PromptSRC & \underline{42.73} & \textbf{37.87} & \textbf{40.15} \\
    \midrule
    \rowcolor{mygray}  
    Ours & 	\textbf{42.97} & \underline{37.07 }& \underline{39.80}  \\
    \bottomrule
    \end{tabular}
    \end{subtable}
    ~
    \begin{subtable}[t]{.32\textwidth}
    \centering
    
    \caption{SUN397.}
    
    \begin{tabular}{l cc|c}
    \toprule
    Method & Base & Novel & HM \\
    \midrule
    CLIP & 69.36 & 75.35 & 72.23 \\
    CoOp & {80.60} & 65.89 & 72.51 \\
    CoCoOp & 79.74 & {76.86} & {78.27} \\
    MaPLe & 80.82 & \underline{78.70} & {79.75} \\
    ProGrad &{81.26} & 74.17 & 77.55 \\
    KgCoOp & 80.29 & 76.53 & 78.36\\
    PromptSRC &  \underline{82.67} & 78.47 & \underline{80.52} \\
    \midrule
    \rowcolor{mygray} 
    Ours & \textbf{82.77} & \textbf{79.50} & \textbf{81.10} \\
    \bottomrule
    \end{tabular}
    \end{subtable}
    
    \begin{subtable}[t]{.32\textwidth}
    \centering
    
    \caption{DTD.}
    
    \begin{tabular}{l cc|c}
    \toprule
    Method & Base & Novel & HM \\
    \midrule
    CLIP & 53.24 & {59.90} & 56.37 \\
    CoOp & {79.44} & 41.18 & 54.24 \\
    CoCoOp & 77.01 & 56.00 & {64.85} \\
    MaPLe & {80.36} & 59.18 & {68.16} \\
    ProGrad & 77.35 & 52.35 & 62.45 \\
    KgCoOp & 77.55 & 54.99 & 64.35 \\
    PromptSRC & \underline{83.37} & \underline{62.97} & \underline{71.75} \\
    \midrule
   \rowcolor{mygray}      
    Ours & \textbf{84.87} & \textbf{65.03} & \textbf{73.64} \\
    \bottomrule
    \end{tabular}
    \end{subtable}
    ~
    \begin{subtable}[t]{.32\textwidth}
    \centering
    
    \caption{EuroSAT.}
    
    \begin{tabular}{l cc|c}
    \toprule
    Method & Base & Novel & HM \\
    \midrule
    CLIP & 56.48 & {64.05} & 60.03 \\
    CoOp & {92.19} & 54.74 & 68.69 \\
    CoCoOp & 87.49 & 60.04 & {71.21} \\
    MaPLe & \textbf{94.07} & {73.23} & \underline{82.35} \\
    ProGrad & 90.11 & 60.89 & 72.67 \\
    KgCoOp & 85.64 & 64.34 & 73.48 \\
    PromptSRC & 92.90 & \underline{73.90} & 82.32 \\
    \midrule
     \rowcolor{mygray}    
    Ours & \underline{93.87} & \textbf{81.27} & \textbf{87.12} \\
    \bottomrule
    \end{tabular}
    \end{subtable}
    ~
    \begin{subtable}[t]{.32\textwidth}
    \centering
    
    \caption{UCF101.}
    
    \begin{tabular}{l cc|c}
    \toprule
    Method & Base & Novel & HM \\
    \midrule
    CLIP & 70.53 & {77.50} & 73.85 \\
    CoOp & {84.69} & 56.05 & 67.46 \\
    CoCoOp & 82.33 & 73.45 & {77.64} \\
    MaPLe & 83.00 & {78.66} & {80.77} \\
    ProGrad & 84.33 & 74.94 & 79.35 \\
    KgCoOp & 82.89 & 76.67 & 79.65 \\
    PromptSRC & \underline{87.10} & \underline{78.80} &  \underline{82.74} \\
    \midrule
    \rowcolor{mygray}     
    Ours &	\textbf{87.50} & \textbf{81.63} & \textbf{84.46}  \\
    \bottomrule
    \end{tabular}
    \end{subtable}
    \label{tab:b2n}}
\end{table*}
\section{Experiments}

In this section, we first introduce the experimental settings. Then, we will present the comparative results of our method with other baselines under these experimental settings. In addition, an ablation study is conducted to analyze the impact of each model component and hyperparameters.

\subsection{Experiment settings}

We extensively evaluate our approach and present a comparison with other methods on three benchmark settings. \\
\textbf{Base-to-novel class generalization.} In this setting, the datasets are split into base and novel classes. The model is trained on base classes with few-shot images and evaluated on both base and novel classes. This benchmark assesses the generalization capability in zero-shot scenarios within a dataset.\\
\textbf{Cross-dataset evaluation.} Following previous work, the model is trained in a few-shot setting on ImageNet \cite{deng2009imagenet} classes and then directly evaluated on ten unseen datasets, without any fine-tuning specific to those datasets.\\
\textbf{Domain generalization setting.} The model is first trained on the ImageNet \cite{deng2009imagenet} dataset and subsequently evaluated on out-of-distribution benchmarks to evaluate its robustness to domain shifts.\\
\textbf{Datasets.} For evaluating both base-to-novel class generalization and cross-dataset performance, we follow CoOp and CoCoOp, utilizing 11 diverse image recognition datasets. These datasets span a wide range of tasks, including generic object recognition with ImageNet~\cite{deng2009imagenet} and Caltech101~\cite{fei2004learning}, fine-grained classification tasks with OxfordPets~\cite{parkhi2012cats}, StanfordCars~\cite{krause20133d}, Flowers102 ~\cite{nilsback2008automated}, Food101~\cite{bossard2014food}, and FGVCAircraft~\cite{maji2013fine}, scene recognition with SUN397~\cite{xiao2010sun}, action recognition with UCF101~\cite{soomro2012ucf101}, texture classification with DTD~\cite{cimpoi2014describing}, and satellite image classification with EuroSAT~\cite{helber2019eurosat}. For domain generalization, ImageNet serves as the source dataset, while ImageNetA~\cite{hendrycks2021natural}, ImageNet-R~\cite{hendrycks2021many}, ImageNet-Sketch~\cite{wang2019learning}, and ImageNetV2~\cite{recht2019imagenet} are used as out-of-distribution datasets.\\
\textbf{Implementation Details.} To ensure a fair comparison, We use CLIP with ViT-B/16 vision encoder across all three benchmark tasks and report results averaged over 3 runs. For base-to-novel setting, we conduct training for 15 epochs and set the learning rate at 0.0035. For domain generalization and cross-dataset setting, we conduct training for 10 epochs and set the learning rate at 0.0025. The size of the learnable prompt pool $M$ set to 12, with each prompt having a length of $L_p=4$. The number of clusters $k$ for image features is set to 4. Additionally, learnable prompt are added to every layer of the text encoder. We set $\lambda_1=0.5$, $\lambda_2=25$, $\lambda_3=0.1$ and $\lambda_4=0.01$ to weight $\mathcal{L}_{align}$, $\mathcal{L}_{cc}$, $\mathcal{L}_{div}$ and $\mathcal{L}_{match}$. These hyperparameters are kept consistent across all three experimental setups. We employ the SGD optimizer with a cosine annealing scheduler on all settings.

\subsection{Base-to-Novel class Generalization}

In this section, we present the quantitative results of various methods on base-to-novel class tasks across 11 datasets. Table \ref{tab:b2n} provides a comprehensive comparison of our method with CLIP, CoOp , CoCoOp, ProGrad, KgCoOp, MaPLe, and PromptSRC. We have highlighted the best results in bold. With the exception of MaPLe \cite{khattak2023maple} and PromptSRC \cite{khattak2023self}, other methods enhance CLIP's performance on base categories but at the expense of generalization ability. This trade-off is evident in their inferior performance on novel categories relative to the original CLIP model. In comparison, MaPLE \cite{khattak2023maple} and PromptSRC \cite{khattak2023self} achieves a balanced performance across both base and novel classes. Remarkably, our method surpasses existing approaches in performance on base categories, novel categories, and the harmonic mean of both. Our method exhibits strong zero-shot generalization, achieving a 2.49\% improvement on novel categories over MaPLe \cite{khattak2023maple} and a 1.53\% improvement over PromptSRC \cite{khattak2023self}. 

For the performance of individual datasets, our method demonstrates particularly significant performance improvements on the novel classes of the DTD \cite{cimpoi2014describing}, EuroSAT \cite{helber2019eurosat}, and UCF101 \cite{soomro2012ucf101} datasets. Specifically, on the DTD dataset \cite{cimpoi2014describing}, our method improves by 2.06\%, achieving 65.03\%; on the EuroSAT dataset , performance improves by 7.37\%, reaching 81.27\%; and on the UCF101 dataset \cite{soomro2012ucf101}, the improvement is 2.83\%, reaching 81.86\%. By leveraging rich and fine-grained visual concepts extracted from the vision encoder, the model enhances cross-modal alignment and effectively transfers knowledge from base categories to unseen categories, thereby improving performance on novel classes.

\subsection{Cross Dataset Evaluation}

\begin{table*}[]
    \caption{ Performance of AttriPrompt on cross-dataset evaluation and its comparison to existing methods. The best results are in bold and the second-best results are underlined.
    }
    \tabstyle{4pt}
    \scalebox{1.0}{
    \begin{tabular}{l|c|ccccccccccc}
    \toprule
    & \textbf{Source} & \multicolumn{11}{c}{\textbf{Target}} \\ \cmidrule(lr){2-2} \cmidrule(lr){3-13}
    & \rotbox{ImageNet} & \rotbox{Caltech101} & \rotbox{OxfordPets} & \rotbox{StanfordCars} & \rotbox{Flowers102} & \rotbox{Food101} & \rotbox{Aircraft} & \rotbox{SUN397} & \rotbox{DTD} & \rotbox{EuroSAT} & \rotbox{UCF101} & \rotbox{\emph{Average}} \\
    \midrule
    CoOp & 71.51 & 93.70 & 89.14 & 64.51 & 68.71 & 85.30 & 18.47 & 64.15 & 41.92 & 46.39 & 66.55 & 63.88 \\
    Co-CoOp & 71.02 & \textbf{94.43} & 90.14 & 65.32 & \underline{71.88} & 86.06 & 22.94 & \underline{67.36} & 45.73 & 45.37 & 68.21 & 65.74 \\
    MaPLe & 70.72 & 93.53 & 90.49 & 65.57 & \textbf{72.23} & \underline{86.20} & \textbf{24.74} & 67.01 & 46.49 & \underline{48.06} & 68.69 & \underline{66.30} \\
    PromptSRC & 71.27 & 93.60 & 90.25 & \textbf{65.70} & 70.25 & 86.15 & 23.90 & 67.10 & \underline{46.87} & 45.50 & \underline{68.75} & 65.81 \\
    \midrule
    \rowcolor{mygray} Ours & \textbf{72.40} & \underline{94.23} & \textbf{90.73} & \underline{65.63} & 71.57 & \textbf{86.47} & \underline{24.37} & \textbf{68.00} & \textbf{48.37} & \textbf{53.17} & \textbf{69.17} & \textbf{67.17}  \\
    \bottomrule
    \end{tabular}}
    \label{tab:xd}
\end{table*}

We compare our cross-dataset performance with previous methods in Table \ref{tab:xd}. The cross-dataset generalization setting assesses whether a model trained on the source dataset can generalize to unseen target datasets, where a distribution shift occurs between base and novel tasks. Our method marks the best performance on the source datasets with the accuracy of 72.4\%. On the target datasets, compared to PromptSRC \cite{khattak2023self}, our method demonstrates competitive performance and achieves better generalization in 9 out of 10 cases, attaining an average of 67.17\% on the target. Notably, our method outperforms MaPLE \cite{khattak2023maple} by 5.11\% on EuroSAT \cite{helber2019eurosat}, a satellite image dataset with distinct characteristics compared to ImageNet \cite{deng2009imagenet}. This indicates that our method effectively generalizes learned visual attribute concepts across datasets.

\subsection{Domain Generalization Experiments}

\begin{table}[!t]
    \caption{Comparison with other methods on domain generalization. The best results are in bold and the second-best results are underlined.} 
    \small \centering
 \setlength{\tabcolsep}{8pt}
    \scalebox{0.8}[0.8]{
    \begin{tabular}{l cccccc}
    \toprule
    & \textbf{Source} & \multicolumn{5}{c}{\textbf{Target}} \\ \cmidrule(lr){2-2} \cmidrule(lr){3-7}
     & ImageNet & -V2 & -S & -A & -R  & Avg.\\
    \midrule
    CLIP &  66.73 & 60.83 & {46.15} & 47.77 & {73.96} & {57.18} \\
    CoOp &  71.51 & {64.20} & 47.99  & 49.71  & 75.21  & {59.28} \\
    Co-CoOp & 71.02 & {64.07} & 48.75 & 50.63 & 76.18 & {59.91}  \\
    ProGrad & \underline{72.24} & \underline{64.73} & 47.61 & 49.39 & 74.58 & 59.07 \\
    Bayesian Prompt & 70.93 & 64.23 & 49.20 & 51.33 & 77.00 & 60.44 \\
    MaPLe & 70.72  & {64.07} & 49.15  & \underline{50.90} & 76.98 & {60.27}  \\
    PromptSRC & 71.27 & 64.35 & \underline{49.55} & \underline{50.90} & \underline{77.80} & \underline{60.65} \\
    \midrule
    \rowcolor{mygray} Ours & \textbf{72.40} & \textbf{65.07} & \textbf{49.6} & \textbf{52.07}  & \textbf{78.17} & \textbf{61.23} \\
    \bottomrule
    \end{tabular}}
        
    \label{tab:dg}
    \vspace{-1em}
\end{table}

We present the results for domain generalization in Table \ref{tab:dg}. Our method exhibits superior performance on both the source and target datasets.  Through a comprehensive evaluation of both source and target datasets, our method demonstrates superior performance and strong cross-domain adaptability. Specifically, on the ImageNet dataset, our method achieves an accuracy of 72.4\%, with the improvement of 1.13\%,  indicating our method's strong ability to adapt to downstream data. Additionally, on the target datasets, our method also shows robust generalization capabilities, outperforming the existing PromptSRC \cite{khattak2023self} method across all datasets, with an average accuracy of 61.24\%. This demonstrates its exceptional performance in cross-domain transfer learning. These results clearly show that our proposed method not only achieves high accuracy on a single downstream dataset but also ensures excellent generalization performance across multiple domains, thereby enhancing the model's robustness and adaptability in real-world applications.

\subsection{Ablation study}

\begin{table}[]
\caption{Ablation study on each component. AR denotes the Attributes Retrieval module.}
    \centering
\begin{tabular}{ccccccc}
\toprule 
 \multirow{2}{*}{AR} & \multirow{2}{*}{ $\mathcal{L}_{div}$ } & \multirow{2}{*}{ $\mathcal{L}_{cc}$ } & \multirow{2}{*}{ $\mathcal{L}_{align}$ } & \multicolumn{3}{c}{ 11 datasets } \\
 \cmidrule{5-7}
 & & & & Base & Novel & HM  \\
\midrule
              &   &  &  & 81.59 & 73.31 & 76.98 \\
  \checkmark  &   &  &  & 83.49 & 75.97 & 79.56  \\
  \checkmark  &  \checkmark  &  &  & 83.52 & 76.26 & 79.73  \\
  \checkmark  &  \checkmark &  \checkmark &  & 83.65 & 77.22 & 80.31  \\
  \rowcolor{mygray}
  \checkmark  &  \checkmark  &  \checkmark & \checkmark & 84.88 & 77.63 & 81.09  \\

\bottomrule

\end{tabular}
    
    \label{tab:each}
\end{table}
\begin{figure}
	\centering
	\includegraphics[width=\columnwidth]{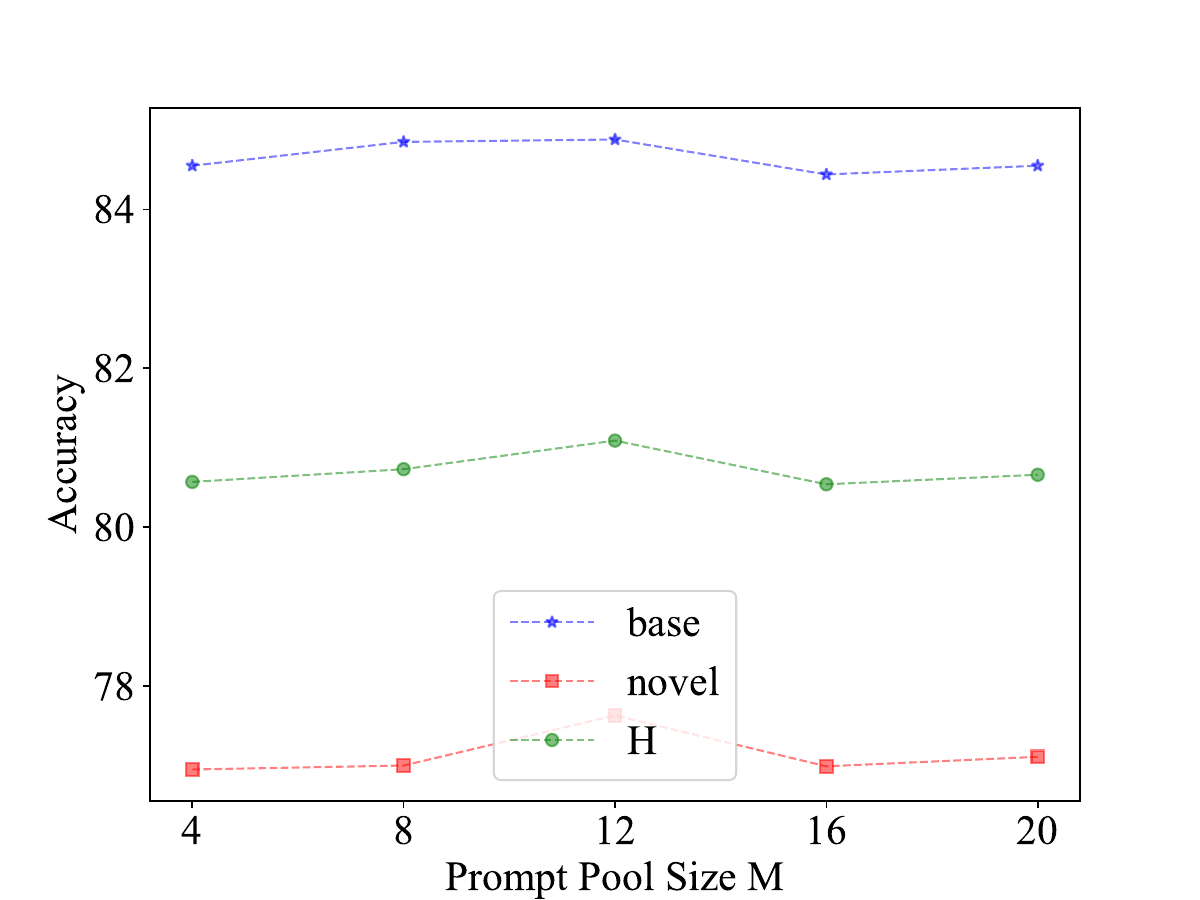} 
	\caption{Ablation study on prompt pool size.}
	\label{fig:abl-prompt-size}
\end{figure}

In this section, we discuss the ablation study, focusing on the the effects of each component and the hyperparameters, including $M$ and $k$. 

To verify the effectiveness of our proposed method, we conducted ablation studies on the key components, with the results summarized in Table \ref{tab:each}. First, the proposed Attribute Retrieval module significantly improves the baseline performance on both base and novel categories. The performance gain can be attributed to the operation of enriching text features with visual information, which promotes better cross-modal alignment and subsequently improves the model’s prediction accuracy. The introduction of orthogonality loss between prompts further enhances the model’s performance on novel categories. This loss promotes semantic diversity within the prompt pool, facilitating generalization to unseen categories. The Self-Regularization loss, by the incorporation of explicit regularization terms that guide the model's gradient updates, substantially improves the model’s generalization capability on unseen categories, reaching 77.22\%. Finally, Dual-stream Contrastive Learning contributes to further improvements in average accuracy by refining and reinforcing the alignment between visual and textual features.

\begin{figure}
	\centering
	\includegraphics[width=\columnwidth]{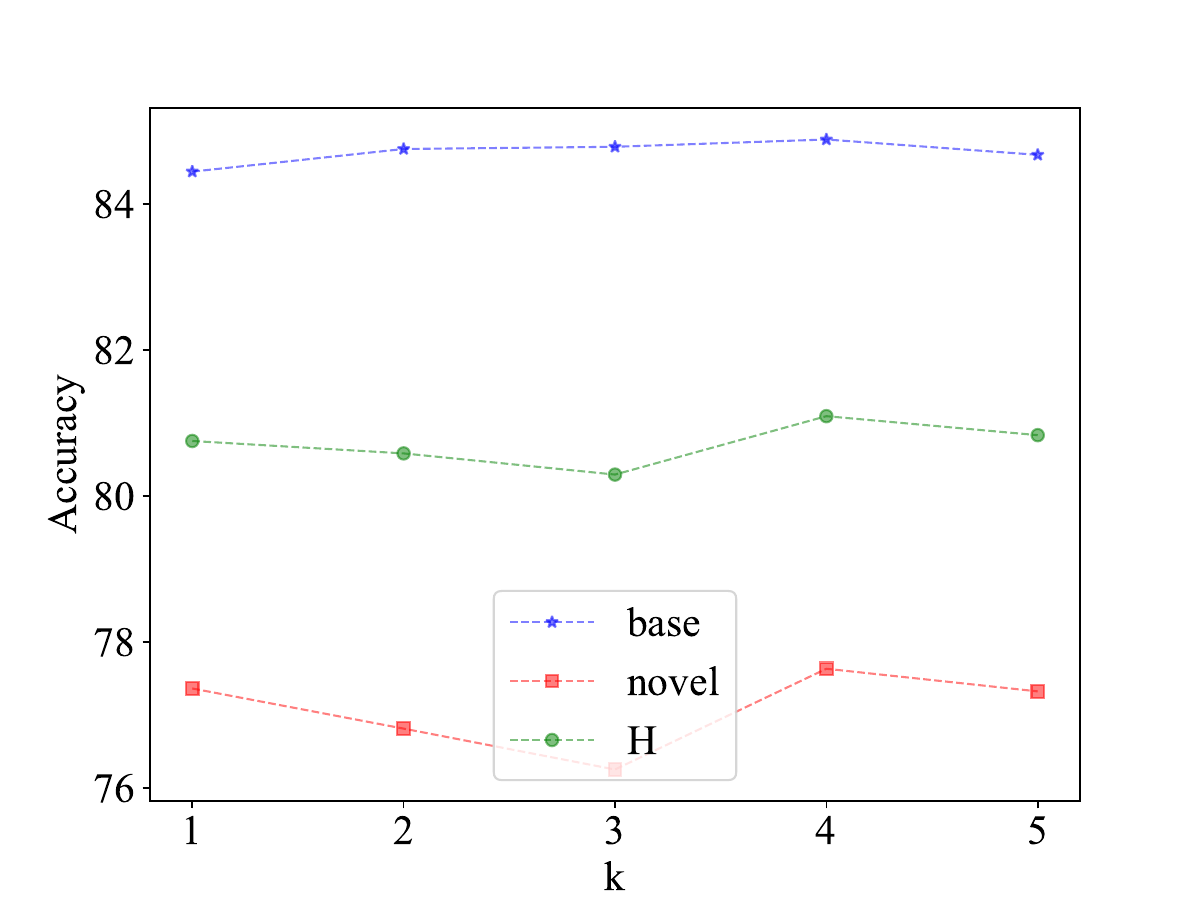} 
	\caption{Ablation study on $k$.}
	\label{fig:abl-k}
\end{figure}

We investigate the effect of the prompt pool size $M$, which controls the number of learnable prompts. As shown in Figure \ref{fig:abl-prompt-size}, when $M$ increases from 4 to 12, the model's performance improves on both base and novel classes. However, beyond a certain threshold, performance begins to degrade. When the size of prompt pool becomes too large, it may introduce a significant number of similar or redundant prompts, thereby increasing the model's susceptibility to irrelevant or suboptimal prompt embeddings and ultimately degrading matching performance.

\begin{figure}
	\centering
	\includegraphics[width=\columnwidth]{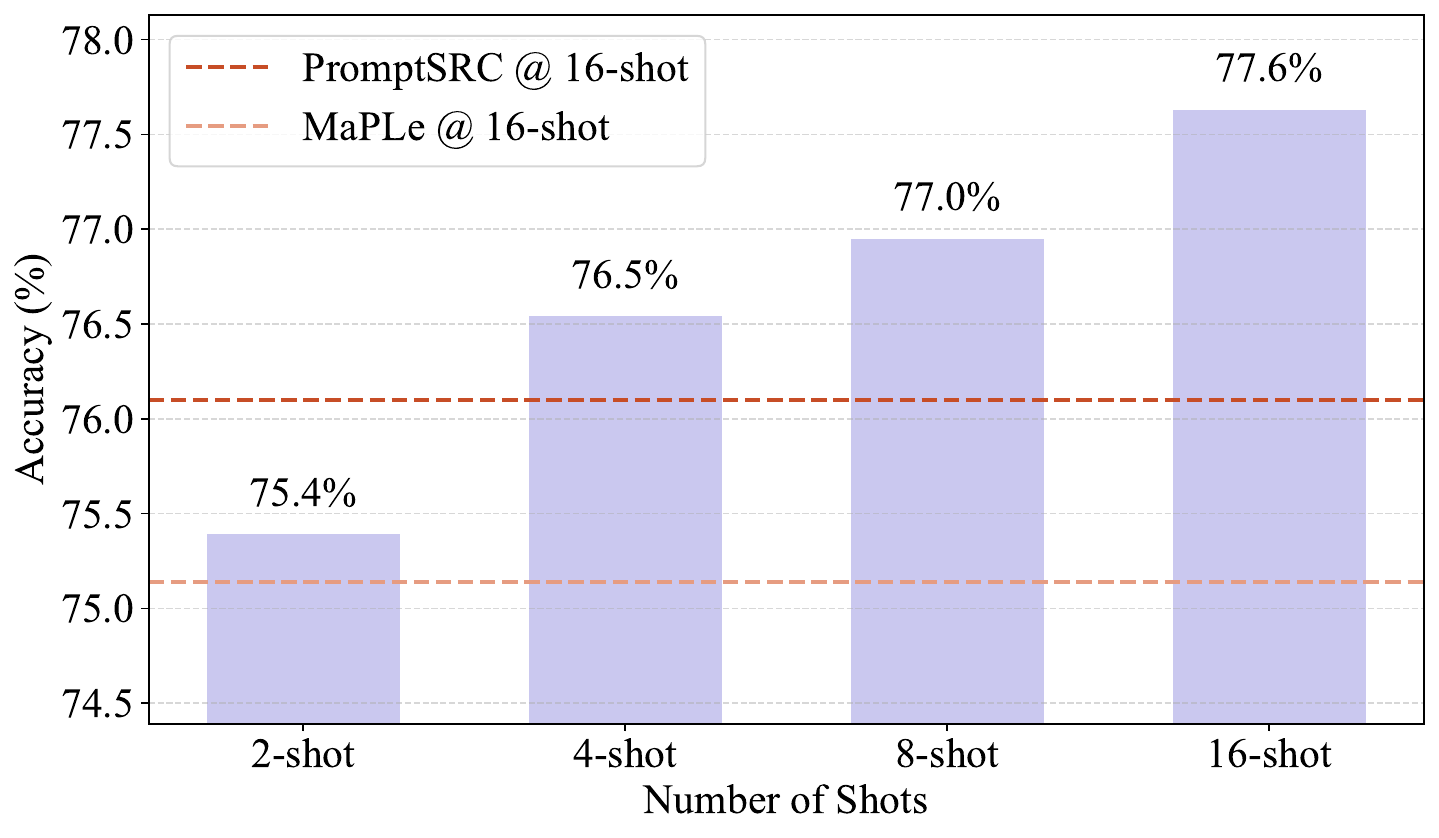} 
	\caption{Comparison of few-shot image recognition among existing methods on 11 datasets.}
	\label{fig:abl-fewshot}
\end{figure}

To better understand the effects of $k$, which controls the number of aggregate attribute features, we conduct an ablation experiment to investigate it. As shown in Figure \ref{fig:abl-k}, when $k = 2$ or $3$, the model exhibits inferior performance on novel classes, possibly due to insufficient representativeness of the clustered features, which leads to suboptimal generalization to novel categories. When $k = 4$, the model achieves the best performance, as the clustered visual features provide adequate coverage of both base and novel classes. However, as $k$ continues to increase, the model's performance begins to decline because of the introduction of noisy features.

Furthermore, we conducted base-to-novel experiments in a few-shot setting with limited data, and the results are shown in Figure \ref{fig:abl-fewshot}. In the 2-shot setting, the model's accuracy is comparable to that of the MaPLe \cite{khattak2023maple} method in the 16-shot setting, resulting in an 8-fold improvement in data efficiency. Compared to the results of PromptSRC \cite{khattak2023self} in the 16-shot setting, our method achieves the same performance in the 4-shot setting, showing a 4-fold improvement in data efficiency. This demonstrates that our method can perform more effectively in knowledge transfer and model adaptation under few-shot conditions, maintaining high performance and significantly reducing data requirements, thereby improving learning efficiency and practical application value.

\section{Conclusion}

In this paper, we propose a prompt tunning method named AttriPrompt. This method not only leverages the attribute information embedded in the visual features from the intermediate layers of the vision encoder to enhance text prompts, but is also capable of adaptively selecting and adjusting prompts based on different input images. AttriPrompt consists of three core components: Attribute Retrieval, Self-Regularization, and Dual-Stream Contrastive Learning. Attribute Retrieval aggregates visual features and retrieves related prompts from prompt pool. Self-Regularization is designed to impose explicit regularization constraints that guide the model’s optimization, regulate its learning trajectory, and mitigate overfitting to specific downstream task distributions. Dual-Stream Contrastive Learning employs a dual-stream contrastive mechanism to perform a fine-grained alignment between vision and text. This helps the model attend to previously overlooked fine-grained features. Extensive experiments demonstrate the efficacy of our method. We believe this may offer new insights and advancements for the field of efficient transfer learning in VLPMs.


\bibliographystyle{ACM-Reference-Format}
\bibliography{sample-base}

\appendix

\section{More Implementation details}

In this section, we provide further implementation details of the proposed approaches presented in the main paper.\\
\textbf{Additional Training details.} We use a publically available ViT-B/16 CLIP model with $d$ = 512 and the batch size is set to 4. The fixed prompts are initialized with the prompt 'a photo of <class>'. The prompts in the prompt pool with the keys are all initialized randomly. We used k-means algorithm with 50 iterations. For base-to-novel setting, we conduct training for 15 epochs and set the learning rate at 0.0035. For domain generalization and cross-dataset setting, we conduct training for 10 epochs and set the learning rate at 0.0025. All models are trained using SGD optimizer and utilize a single NVIDIA A6000 GPU.\\
\textbf{Details of comparing methods.} We performed a thorough comparison between our method and others approaches. To guarantee a fair comparison, all methods were implemented using the same version of CLIP, and their experimental setups adhered to the hyperparameters outlined in their papers. The compared methods are as follows: 

\begin{itemize}\setlength{\itemsep}{0em}

\item CoOP \cite{zhou2022learning}, the first method to employ prompt learning in CLIP-like vision-language models. 

\item CoCoOP \cite{zhou2022conditional}, Which improves CoOp by adding a meta-network, thereby enhancing the generalization ability during the fine-tuning process. 

\item KgCoOP \cite{yao2023visual}, by utilizing the knowledge embedded in the original CLIP, the model’s generalization capability is preserved.

\item ProGrad \cite{zhu2023prompt},  data-driven method which learns low-bias prompts from a few samples.

\item Bayesian Prompt \cite{derakhshani2023bayesian}, a method that addresses the fine-tuning problem from a Bayesian perspective, defining it as a variational inference problem.

\item MaPLe \cite{khattak2023maple}, the first fine-tuning method based on multimodal prompts.

\item PromptSRC \cite{khattak2023self}, refines learned prompts by leveraging pre-trained features to improve generalization capability.

\end{itemize} 
\textbf{Evaluation metrics.} In the base-to-novel generalization setting, we evaluate performance by reporting the top-1 accuracy on both base and novel classes across all datasets. To quantify the overall generalization ability, we also compute the harmonic mean (HM) of the two accuracies, which serves as the key metric. For cross dataset evaluation and domain generalization experiments, we report top-1 accuracies obtained on the corresponding test set of each dataset using the splits provided in CoOp.

\section{Compute Cost analysis}

In the attribute retrieval stage, given $k$ clustered image features $a \in \mathbb{R}^{k\times d}$ and $m$ prompt keys $keys \in \mathbb{R}^{m \times d}$, the retrieval has a time complexity of $O(mnk)$. Since  $m$ and $k$ are typically small, the computational overhead is limited. Similarly, the dual-stream contrastive learning module also introduces minimal computational overhead.
For the visual clustering stage, applying k-means to $n\times n$ image features $x \in \mathbb{R}^{n\times n\times d}$ yields a time complexity of $O(n^2d)$. This is comparable to a single attention operation in ViT and therefore only slightly increases the overall computation.

We evaluate computational cost on DTD using a single 3090 over 15 epochs. Compared to PromptSRC in Table \ref{tab:method_comparison}, AttriPrompt improves performance by 4.94\% with minor increases in training time (+8.4\%), testing time (+4.7\%), and computation (+0.09 GFLOPs), indicating superior efficiency and performance.

\begin{table}[htbp]
\caption{Comparison of methods in terms of training time, testing time and GFLOPs.}
\centering
\begin{tabular}{lcccc}
\toprule
\textbf{Method} & \textbf{Train (min)} & \textbf{Test (min)} & \textbf{GFLOPs} & \textbf{HM} \\
\midrule
PromptSRC    & 2.36 & 0.19 & 89.58 & 68.70 \\
\rowcolor{mygray} AttriPrompt  & 2.56 & 0.20 & 98.32 & 73.64 \\
\bottomrule
\end{tabular}
\label{tab:method_comparison}
\end{table}

\section{More Ablation study} 

\begin{figure}
	\centering
	\includegraphics[width=\columnwidth]{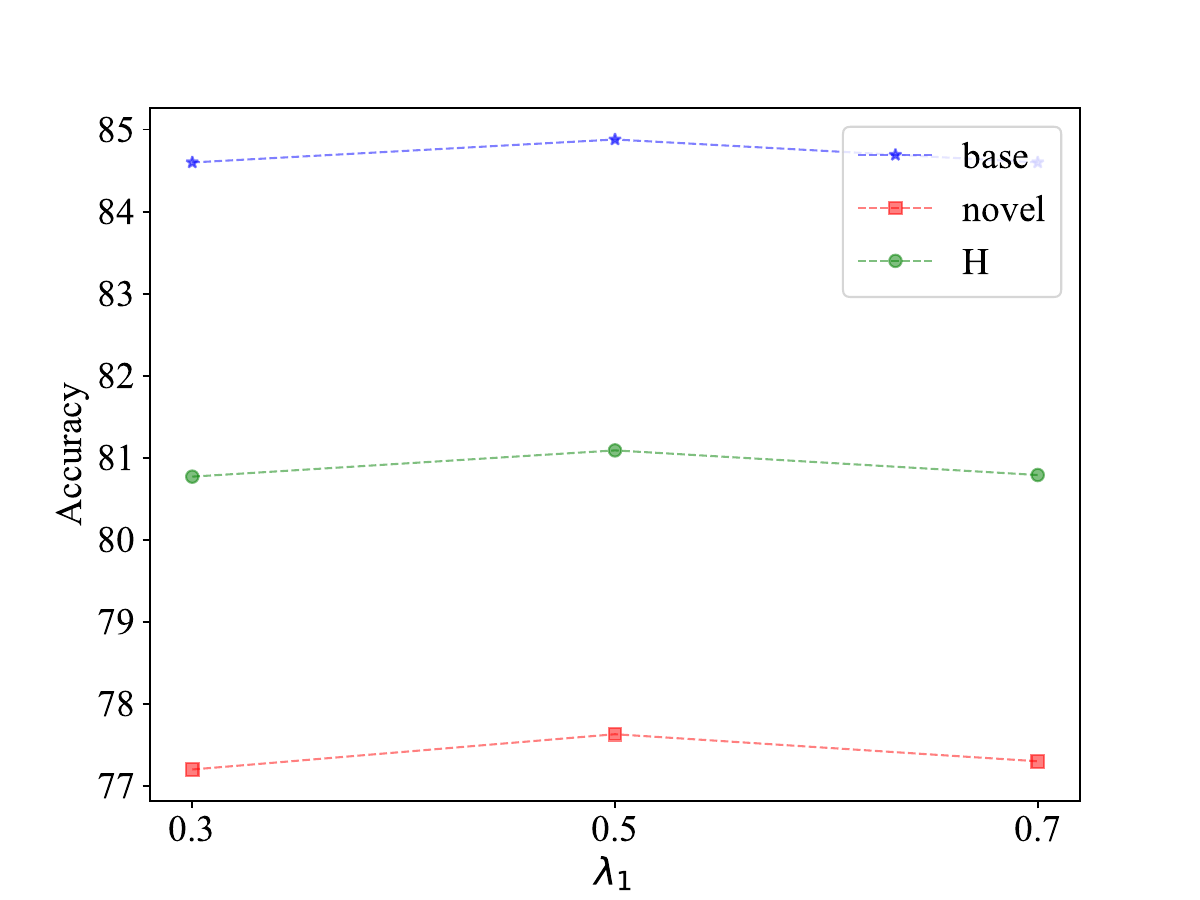} 
	\caption{Ablation study on $\lambda_1$.}
	\label{fig:abl-lambda}
\end{figure}

\begin{figure}
	\centering
	\includegraphics[width=\columnwidth]{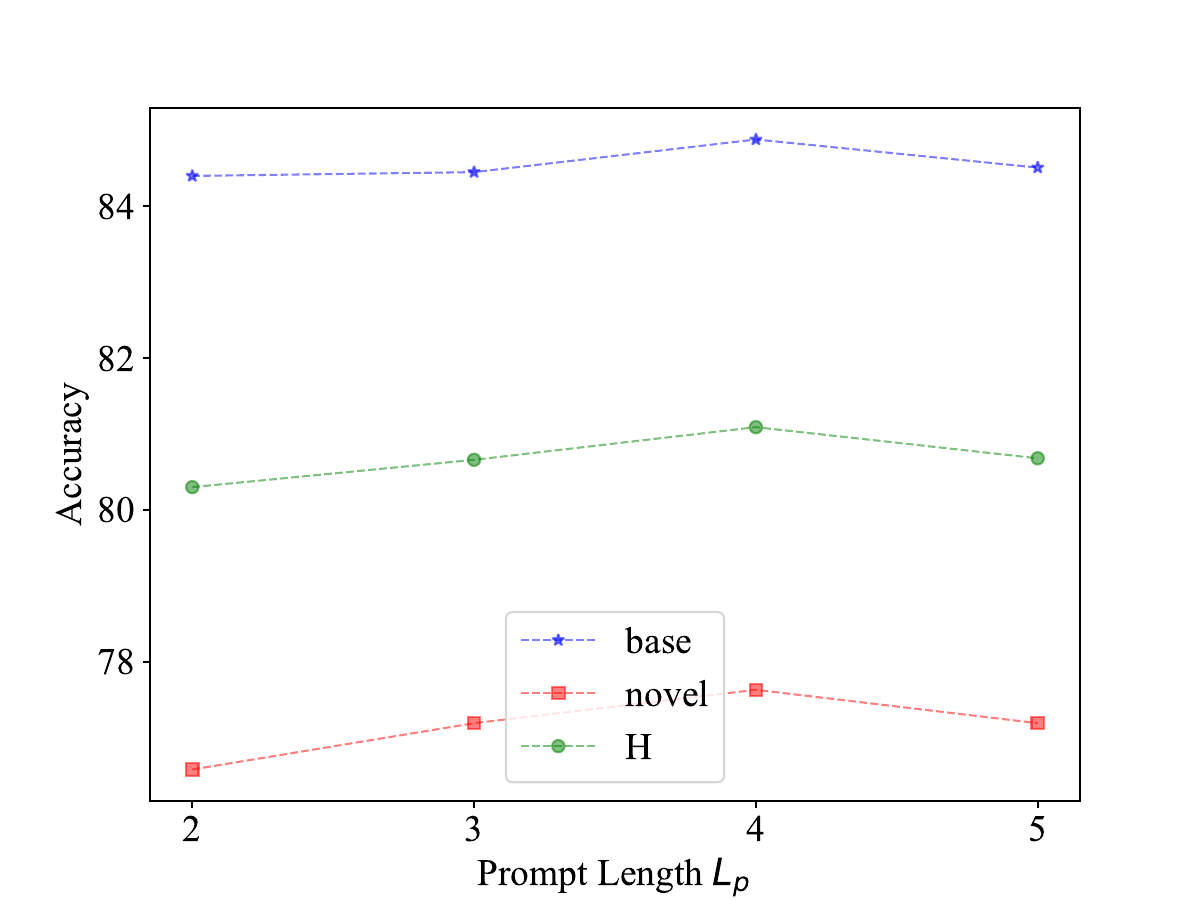} 
	\caption{Ablation study on $L_p$.}
	\label{fig:abl-length}
\end{figure}

\textbf{Dual-stream Contrastive Learning hyper-parameters.} Since the hyperparameter $\lambda_1$ can influence the model's predictions during inference, we conduct an ablation study to evaluate its impact. As shown in Figure \ref{fig:abl-lambda}, the model achieves optimal performance when $\lambda_1 = 0.5$, while slight performance degradation is observed at $\lambda_1 = 0.3$ or $0.7$. When $\lambda_1 = 0.3$, the model relies more on the pre-trained alignment and fails to fully exploit the newly introduced prompts. In contrast, when $\lambda_1 = 0.7$, the new features are overly activated, causing the model to overemphasize the text prompt features and thereby disrupting the alignment learned during pre-training.\\
\textbf{Prompts Length $L_p$.} To better understand the effects of $L_p$, which controls the length of the learnable prompts, we conduct an ablation experiment on it. As shown in Figure \ref{fig:abl-length}, when $L_p = 2$ or $3$,  the model underperforms on both base and novel classes compared to the optimal configuration. This may be attributed to the insufficient length of the prompt, which limits the amount of visual information that can be integrated into the text features, thereby hindering effective cross-modal representation learning. The best performance is achieved when $L_p = 4$, where the prompt provides sufficient parameter capacity to encode both low-level visual attributes and high-level semantics. However, further increasing the prompt length results in performance degradation, possibly due to the increased number of parameters requiring more training data to fully converge. In few-shot scenarios, this can lead to suboptimal convergence and local minima. \\
\textbf{Sensitivity Analysis of $\lambda_3$ and $\lambda_4$.} We further analyzed the sensitivity of $\lambda_3$ and $\lambda_4$ via grid search on two representative datasets: Caltech101 and DTD datasets. As shown in Table \ref{tab:lambda_ratio} , the average HM score across these datasets shows negligible variation, suggesting that AttriPrompt demonstrates strong robustness to changes in these hyperparameters. However, setting either of them to 0 would result in a significant drop in model performance. Accordingly, we set $\lambda_3 = 0.1$ and $\lambda_4 = 0.01$ for all experiments. 

\begin{table}[htbp]
\caption{Sensitivity analysis results of $\lambda_3$ and $\lambda_4$.}
\centering
\begin{tabular}{ccccc}
\toprule
\diagbox{$\lambda_3$}{$\lambda_4$} & \textbf{0} & \textbf{0.01} & \textbf{0.04} & \textbf{0.07} \\
\midrule
0   & 81.87 & 84.11 & 84.03 & 84.21 \\
0.1 & 82.94 & 85.31 & 85.23 & 85.28 \\
0.4 & 82.50 & 85.26 & 85.21 & 85.25 \\
0.7 & 82.63 & 85.26 & 85.21 & 85.37 \\
\bottomrule
\end{tabular}
\label{tab:lambda_ratio}
\end{table}

\end{document}